\documentclass[letterpaper]{article} 
\usepackage{aaai23}  
\usepackage{times}  
\usepackage{helvet}  
\usepackage{courier}  
\usepackage[hyphens]{url}  
\usepackage{graphicx} 
\usepackage{natbib}  
\usepackage{caption} 
\usepackage{algorithmic}

\usepackage{amsmath}
\usepackage[ruled,vlined]{algorithm2e}
\usepackage{xcolor}
\usepackage{caption}
\usepackage{enumitem}
\usepackage{adjustbox}
\usepackage{multirow}
\usepackage{mathtools}

\DeclarePairedDelimiter\floor{\lfloor}{\rfloor}
\usepackage{float}

\title{Task-Adaptive Meta-Learning Framework for Advancing Spatial Generalizability}
\author{
    Zhexiong Liu\textsuperscript{\rm 1}, Licheng Liu\textsuperscript{\rm 2}, Yiqun Xie\textsuperscript{\rm 3}, Zhenong Jin\textsuperscript{\rm 2}, Xiaowei Jia\textsuperscript{\rm 1}
}
\affiliations{
    \textsuperscript{\rm 1}University of Pittsburgh, \textsuperscript{\rm 2}University of Minnesota, \textsuperscript{\rm 3}University of Maryland\\
    \{zhexiong.liu, xiaowei\}@pitt.edu, \{lichengl, jinzn\}@umn.edu, xie@umd.edu
}

\usepackage{bibentry}

\begin{document}

\maketitle

\begin{abstract}

Spatio-temporal machine learning is critically needed for a variety of societal applications, such as agricultural monitoring, hydrological forecast, and traffic management. These applications greatly rely on regional features that characterize spatial and temporal differences. However, spatio-temporal data often exhibit complex patterns and significant data variability across different locations. The labels in many real-world applications can also be limited, which makes it difficult to separately train independent models for different locations. Although meta learning has shown promise in model adaptation with small samples, existing meta learning methods remain limited in handling a large number of heterogeneous tasks, e.g., a large number of locations with varying data patterns. To bridge the gap, we propose task-adaptive formulations and a model-agnostic meta-learning framework that ensembles regionally heterogeneous data into location-sensitive meta tasks. We conduct task adaptation following an easy-to-hard task hierarchy in which different meta models are adapted to tasks of different difficulty levels. One major advantage of our proposed method is that it improves the model adaptation to a large number  of heterogeneous tasks. It also enhances the model generalization  by automatically adapting the meta model of the corresponding difficulty level to any new tasks. We demonstrate the superiority of our proposed framework over a diverse set of baselines and state-of-the-art meta-learning frameworks. Our extensive experiments on real crop yield data show the effectiveness of the proposed method in handling spatial-related heterogeneous tasks in real societal applications. 
\end{abstract}
\section{Introduction}

\begin{figure}[t]
    \centering
    \includegraphics[width=0.89\columnwidth]{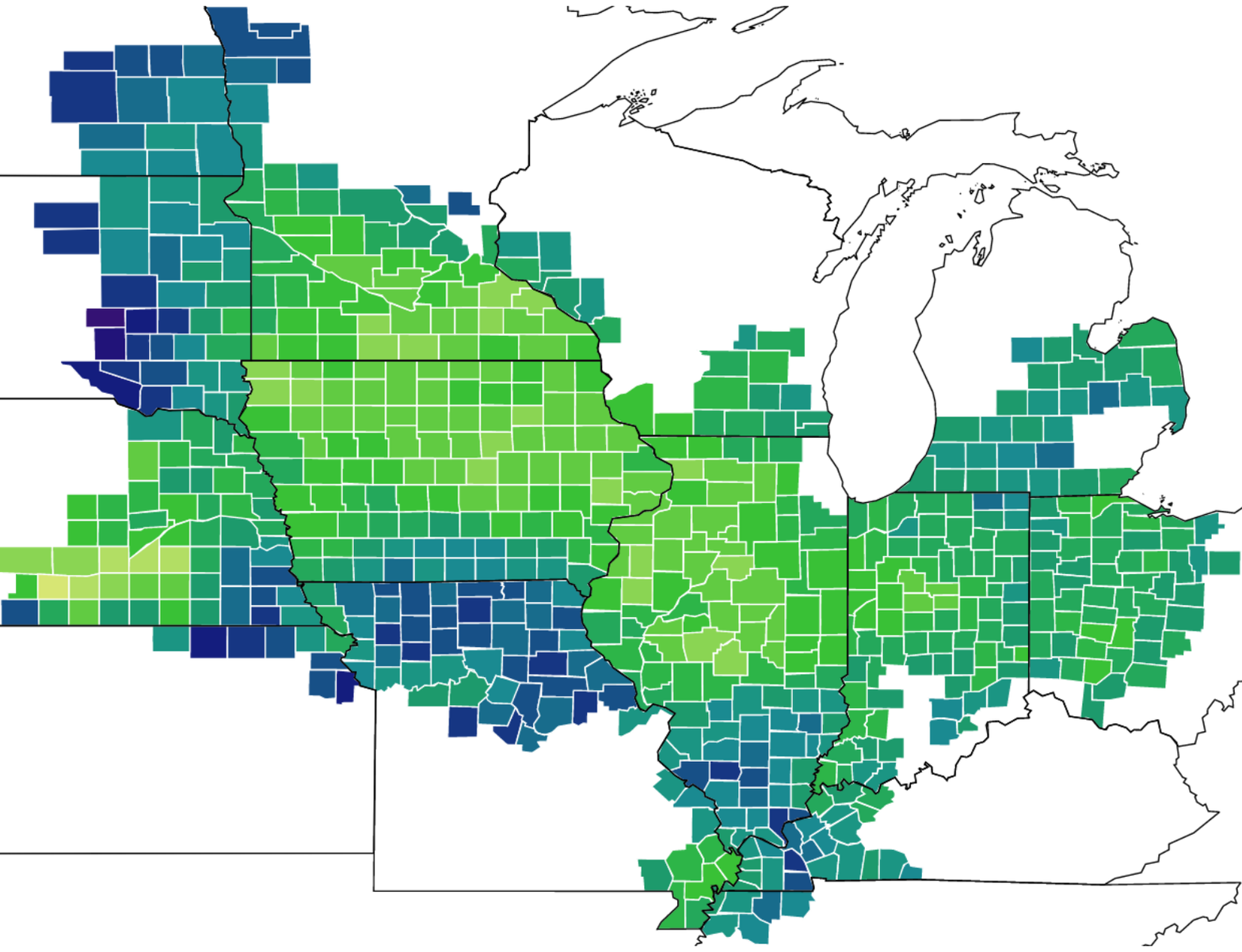}
    \caption{Normalized average corn yields in 21 consecutive years from 2000 to 2020 across 630 counties in Illinois, Wisconsin, Minnesota, Iowa, Missouri, Indiana, Ohio, Michigan, North Dakota, South Dakota, Nebraska, Kansas, Kentucky, and Tennessee in the United States. The dark blue means low-yield regions while light green represents high-yield regions. Geographically, corn yields are highly correlated to spatial locations that are complex to model with contemporary machine learning methods.}
    \vspace{-.2in}
    \label{fig:example}
\end{figure}

The explosive growth of spatio-temporal data emphasizes the needs for automatically discovering spatial-related knowledge \cite{shekhar2003trends}. Spatio-temporal data are complex due to inherent data characteristics such as implicit spatial relationships between variables and the data variability across locations \cite{ huang2018deepcrime, zheng2020gman,huang2020lsgcn}. For example, Figure \ref{fig:example} shows the normalized average corn yield for every county in the Midwestern United States. The yield data exhibit a strong spatial variability due to the variation in weather, soils, and management practices across different counties.
Hence,  a global model trained over large regions may not perform well for every individual county \cite{karpatne2018machine}.
In addition, it is often expensive to collect a large number of labeled samples in real-world societal applications, which makes it challenging to train advanced deep neural network models separately for each location. Therefore, the development of effective machine learning techniques for spatial-related tasks with strong spatial variability is urgently needed.

Transfer learning methods have been widely explored for adapting machine learning models across space. For example, previous works use a Long-Short Term Memory (LSTM) structure with the attention mechanism to transfer spatial-related information~\cite{nigam2019crop, sharma2020wheat, jiang2022learning}. However, these methods directly learn on global data but do not consider regional discrepancy across space. Recent deep learning-based domain adaptation approaches \cite{nevavuori2019crop, elavarasan2020crop} have demonstrated the success on several tasks when trained with sufficiently labeled data; however, their performance can be degraded given limited labeled data in regression tasks.

Few-shot learning has  shown promise in reducing the need for large labeled samples. Nevertheless, standard few-shot learning methods often perform worse if data are from heterogeneous  distributions, which is a common issue in real spatial datasets. Meta learning addresses this issue through the idea of task-adaptive learning. Specifically, meta learning aims to extract meta knowledge from multiple training tasks, which can then be  used to  facilitate task-adaptive learning for a single task using a small number of data samples. Meta learning has shown encouraging results in many important societal problems, such as agricultural monitoring and traffic management~\cite{pan2020spatio, li2020learning, tseng2021learning}.  
Existing meta-learning methods can be 
categorized based on how they leverage meta knowledge in new tasks, 
e.g., 
the optimization-based methods~\cite{finn2017model,li2017meta,antoniou2018train}, the feed-forward model-based methods~\cite{mishra2017simple,qiao2018few}, and metric-learning-based methods~\cite{sung2018learning,willard2021predicting}. 
For example, the Model-Agnostic Meta Learning (MAML) algorithm \cite{finn2017model} aims to learn an initial model (i.e., meta-model) that can be quickly adapted to new tasks. However, most of existing meta-learning methods have limits in handling  a large number of heterogeneous tasks, e.g., modeling data from a large number of locations with non-stationary relationships between input and output variables. This can be a common issue in many societal applications. For example, the variation of weather and soils over space interact with the complex carbon, nitrogen and water cycles during crop growth, which ultimately leads to a strong variability in crop yield patterns.

In this paper, we develop a task-adaptive meta-learning framework by adapting the predictive model gradually over space via a ``spatialized" easy-to-hard task hierarchy. In particular, we first train a standard MAML model by considering each location as a separate task. Then we iteratively split the set of tasks to create new branches of harder tasks. Moreover, we synchronously transform the meta-learning model following the obtained  easy-to-hard task hierarchy. Given a new task, we can first identify its difficulty level and then adapt the meta-model from the corresponding layer of the hierarchy to the new task. Our contributions can be summarized as follows:
\begin{itemize}[leftmargin=*]
  \item We create the first meta-learning method that uses spatial-related tasks in crop yield prediction, which is critical for ensuring food supply and estimating farmers' insurance and subsidies;
  \item  We propose a new meta-learning strategy to learn different difficulty levels of tasks in an easy-to-hard hierarchy that can be quickly adapted to new tasks; 
  \item We extend existing meta-learning methods to handle a large number of heterogeneous tasks; 
  \item Our evaluation on real crop yield data over large regions shows the superiority of our proposed approach over standard machine learning and meta-learning baselines. 
    
\end{itemize} 
\section{Related Work}
\textbf{Few-shot Meta Learning} Few-shot learning has been widely adopted for addressing real-world small data problems due to its great diversity and feasibility \cite{thrun1998learning, finn2017model, wang2020generalizing}. Typically, few-shot learning has gained attention in three fields: (1) metric learning-based methods that learn a similarity space,  which helps build the connections between new few-shot examples with existing data~\cite{vinyals2016matching,snell2017prototypical,jiang2020multi, matsumi2021few}; (2) memory network-based methods that learn to gain experience in training, and generalize learned knowledge to unseen tasks \cite{santoro2016meta, munkhdalai2017meta, zhao2021learning}; (3) Gradient descent-based meta-learning methods that learn to adapt a specific base-learner to  few-shot examples from different tasks. For example, MAML \cite{finn2017model} uses a meta learner to find the optimal initialization for a base learner and adapts it to new
learning tasks with a few training samples. However, existing MAML-based methods have degraded performance given a large number of heterogeneous tasks, such as spatial-related tasks with strong data variability across  different locations. 

\textbf{Multi-task Learning} Multi-task learning (MTL) aims to learn shared representations jointly from multiple training tasks \cite{caruana1997multitask}. It assumes the shared information across different tasks
can be leveraged to improve the overall performance in all tasks \cite{zhang2018overview, ma2020efficient}. These approaches assume that such shared representations could transfer to other tasks, such as object detection \cite{zhang2014facial, li2016deepsaliency}, image segmentation \cite{kendall2018multi}, multi-lingual machine translation \cite{dong2015multi, zhou2019improving} and understanding \cite{liu2019multi, wu2020understanding}. However, the spatial-related tasks cannot be directly learned with the multi-task objectives. This is because location-based data (e.g., crop yields across the United States) have significantly different distributions based on their geographic features. In addition, tasks are relatively independent in real scenario problems thus unable to jointly learn an overall model that benefits every task. A potential solution is to explore task-based feature relations \cite{zhao2020efficient}; however, it requires sufficient labels thus has limitations in many real applications.

\begin{figure*}[t]
    \centering
    \includegraphics[width=0.99\textwidth]{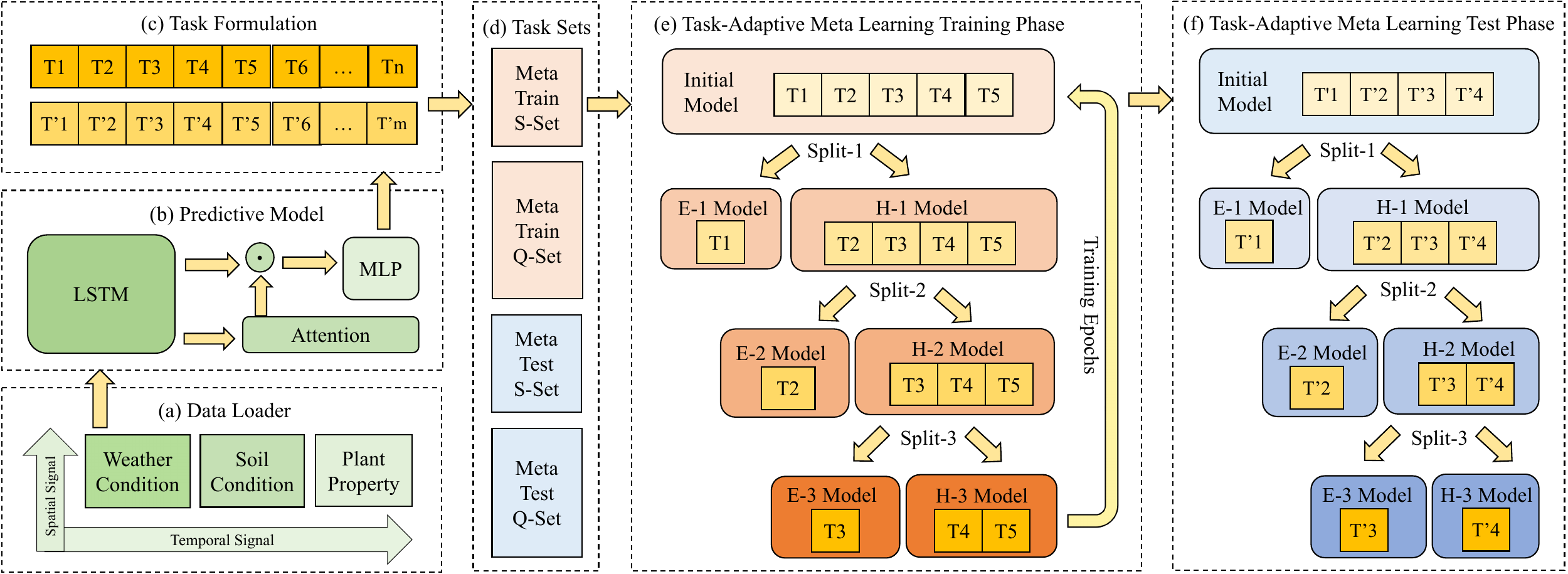}
    \vspace{-.05in}
    \caption{The framework of task-adaptive MAML with an easy-to-hard hierarchy. (a) The weather and soil condition and plant property data that has both spatial and temporal distributions are loaded to (b) train the predictive model. (c) We formulate meta-learning tasks as county-level crop yield predictions of which each meta task is trained with a predictive model. (d) The tasks are split into meta training and meta test sets based on their spatial (county) locations, of which meta training and test sets are split into a support set (S-Set), and a query set (Q-Set) based on their temporal (year) information, respectively. (e) In the adaptively training phase, the Train S-Set is used to train an initial model (that is either a pretrained predictive model or a hard model in the bottom task layer obtained from the last epoch training) and the trained meta model is evaluated on Train Q-Set using metric R$^2$. An array of task-specific R$^2$ is used as the input of Algorithm \ref{algo:gamma} that returns an easy-hard task splitting threshold ($\gamma$). In the first task layer ($r=0$), the task $T_1$ has R$^2$ greater than the threshold Split-1 thus is used to train an easy (E-1) model, otherwise, the tasks ($T_2$ to $T_5$) are used to train a hard (H-1) model. The adaptive training will run multiple splitting iterations until the maximum split number is reached (e.g., 3 splits). The splitting thresholds ($\gamma$) will update in each epoch training as shown in Table \ref{tab:split-threshold}. (f) In the adaptive test phase, input tasks ($T^\prime_1$ to $T^\prime_4$) are split by thresholds ($\gamma$) updated in the training, and apply the same splitting strategies used in training to obtain its best adaptive meta model.} 
    \label{fig:framework}
    \vspace{-.18in}
\end{figure*}
\section{Methods}
\subsection{Preliminaries}
In this section, we introduce notation and  definition used in the crop yield prediction problem. This is essentially a regression task that involves real spatio-temporal data collected in the Midwestern United States. The inputs are an array of daily-collected time-series features, including weather and soil conditions and plant property. The target variable is the county-level  crop yield (for corns) at a yearly scale. Specifically, let $x_i\in \mathcal{R}^{s\times t}$ denote input features and $y_i \in \mathcal{R}$ denote crop yield label of the U.S. county $i$, where $i\in \{1,2 \ldots n\}$, $s$ is the spatial dimension (i.e., the number of locations sampled from each county), and $t$ is the temporal dimension. Let
$T_i=(X_i,Y_i)$
denote the crop yield prediction task
of county $i$ with
the input data
$(X_i,Y_i)=\left\{\left(x_{p}^{(i)}, y_{p}^{(i)}\right);\left(x_{q}^{(i)}, y_{q}^{(i)}\right)\right\}$, $p\in\{1,2,\ldots,k\}$, and $q\in\{1,2\ldots, l\}$.
Here
$\left(x_{p}^{(i)}, y_{p}^{(i)}\right)$ is a task-specific training sample from the Train Support Set (Train S-Set) of size $k$, and $\left(x_{q}^{(i)}, y_{q}^{(i)}\right)$ is a validation sample from the Train Query Set (Train Q-Set) of size $l$, which is reserved for evaluating the training task performance. 

MAML \cite{finn2017model} aims to solve meta-learning problems by optimizing the adaptability of the meta model $\mathcal{F}_{\theta}$. It learns parameter $\theta$ over a set of training tasks $\mathcal{T}_{train}$ where $\mathcal{T}_{train} = \left\{T_1, T_2, T_3, ..., T_n\right\}$,
 such that the learned meta model $\mathcal{F}_{\theta}$ is able to quickly solve new tasks $T^\prime_j \in \mathcal{T}_{test}$ 
by slightly  fine-tuning $\mathcal{F}_{\theta}$ with a small amount of task-specific samples $(X^\prime_j, Y^\prime_j)$. Here  $\mathcal{T}_{test} = \left\{T^\prime_1, T^\prime_2, T^\prime_3, ..., T^\prime_m\right\}$, and
$(X^\prime_j,Y^\prime_j)=\left\{\left({x^\prime}_{p^\prime}^{(j)}, {y^\prime}_{p^\prime}^{(j)}\right);\left({x^\prime}_{q^\prime}^{(j)}, {y^\prime}_{q^\prime}^{(j)}\right)\right\}$, $p^\prime=1,2,\ldots,k^\prime, q^\prime=1,2\ldots, l^\prime$.
Here $\left({x^\prime}_{p^\prime}^{(j)}, {y^\prime}_{p^\prime}^{(j)}\right)$ is a sample used to fine-tune the learned meta model from the Test Support Set (Test S-Set), and $\left({x^\prime}_{q^\prime}^{(j)}, {y^\prime}_{q^\prime}^{(j)}\right)$ is the evaluation sample for the performance in the Test Query Set (Test Q-Set).

Specifically, we train $\mathcal{F}_{\theta}$ on the task $T_{i} \in \mathcal{T}_{train}$ with gradient descent optimization
\begin{equation}
\theta_{i} \leftarrow \theta-\alpha \nabla \mathcal{L}_{T_{i}}\left(\mathcal{F}_{\theta}\right)
\label{eq:theta_i},
\end{equation}
\begin{equation}
    \mathcal{L}_{T_{i}}\left(\mathcal{F}_{\theta}\right)=\frac{1}{k} \sum_{p=1}^{k} \ell\left(\mathcal{F}_{\theta}\left(x_{p}^{(i)}\right), y_{p}^{(i)}\right),
    \label{eq:task_loss}
\end{equation}
\noindent where $\mathcal{L}_{T_{i}}\left(\mathcal{F}_{\theta}\right)$ is the task-related training (outer) loss, $\alpha$ is a meta-learning rate, and $\ell$ is the associated (inner) loss (e.g., mean squared loss, MSE). 
In the adaptation stage, MAML optimizes $\theta$ such that the following meta loss is minimized using the task-wise fine-tuned parameter $\theta_{i}$ over validation samples of each training task 
$\left(x_{q}^{(i)}, y_{q}^{(i)}\right)$:
\vspace{-.1in}
\begin{equation}
\min _{\theta} \mathcal{L}_\text{MAML}\left(\mathcal{F}_{\theta}\right)\!=\!\frac{1}{l}\sum_{i=1}^{l} \ell\left(\mathcal{F}_{\theta-\alpha \nabla_{\theta} \mathcal{L}_{T_{i}}\left(\mathcal{F}_{\theta}\right)}\left(x_{q}^{(i)}\right), y_{q}^{(i)}\right),
\label{eq:theta_loss}
\end{equation}

The meta parameter $\theta$ is then updated by gradient descent
$\theta \leftarrow \theta-\beta \nabla_{\theta} \mathcal{L}_\text{MAML}\left(\mathcal{F}_{\theta}\right)$. The learned meta model $\mathcal{F}_{\theta}$ can be used to fine-tune a new task $T^\prime_j \in \mathcal{T}_{test}$ through Eq. \ref{eq:theta_i}.

\subsection{Predictive Model}
\label{sec: predictive-model}
In this section, we introduce the LSTM-Attention network \cite{xu2020deepcropmapping} for the corp yield prediction, as shown in Figure \ref{fig:framework} (b). The inputs are fed to the LSTM layer to learn hidden states and their attentions. The outputs are model predictions learned by a multi-layer perception. In particular, the LSTM module trains on the spatial and temporal input features to learn its weights by computing multiple gates (input gate $i$, forget gate $f$, and output gate $o$) that determine whether the incoming data stream should retain or forget: 
\begin{equation}
i_{t}=\sigma\left(W_{x i} x_{t}+W_{h i} h_{t-1}+W_{c i} c_{t-1}+b_{i}\right) 
\label{eq:start_predictive_mode}
\end{equation}
\begin{equation}
 f_{t}=\sigma\left(W_{x f} x_{t}+W_{h f} h_{t-1}+W_{c f} c_{t-1}+b_{f}\right) 
\end{equation} 
\begin{equation}
c_{t}=f_{t} c_{t-1}+i_{t} \tanh \left(W_{x c} x_{t}+W_{h c} h_{t-1}+b_{c}\right) 
\end{equation}
\begin{equation}
o_{t}=\sigma\left(W_{x o} x_{t}+W_{h o} h_{t-1}+W_{c o} c_{t}+b_{o}\right) 
\end{equation} 
\begin{equation}
h_{t}=o_{t} \tanh \left(c_{t}\right)
\end{equation}
\noindent where \(\sigma\) denotes the sigmoid function, $c$ is
the cell state, $W$ is the weight matrix, $b$ is the bias term, $h$ is the hidden state, and $t$ is the time step. In addition, we use an attention module that contains several dense layers and a softmax layer to learn the attention $\alpha$ for each hidden state $h_t$ at time step $t$
\begin{equation}
\alpha_{t}=\operatorname{Softmax}\left(W_{ att } \cdot h_{t}+b_{att }\right),
\end{equation}
\noindent where $W_{att}$ and $b_{att}$ are attention weight and bias, respectively. Afterward, the aggregated attention $\alpha$ and hidden state $h$ over all the time steps are fed to a multi-layer perception that returns predicted corp yield $\hat{y}$, as
\begin{equation}
    \hat{y} = \text{MLP}(\alpha \cdot h).
    \label{eq:end_predictive_mode}
\end{equation}

\subsection{Task-adaptive MAML}
Existing meta-learning approaches require 
new testing task 
$T^{\prime} \in \mathcal{T}_{test}$
to be from the same distribution as the training tasks $\mathcal{T}_{train}$. The adaptation performance to the new task $T^{\prime}$ can often be degraded when the training task distribution $p(\mathcal{T}_{train})$ is highly heterogeneous due to a large number of training tasks $\mathcal{T}_{train}$. To address this issue, we consider decomposing the training task distribution $p(\mathcal{T}_{train})$ based on the task difficulty level, and have the model be adapted gradually following an easy-to-hard task hierarchy, as shown in Figure \ref{fig:framework} (e). In particular, we start with building an initial predictive model using all the task samples in the Train S-Set. Different predictive models can be used in the proposed framework thus we adopt the LSTM-Attention network (introduced from Eq. \ref{eq:start_predictive_mode} to \ref{eq:end_predictive_mode}) to train a meta model on the Train S-Set and optimize its MSE loss using Eq. \ref{eq:theta_i}.

The performance of the learned meta model on the validation data of each task $T_i$ from the Train Q-Set can serve as a proxy measure for the task difficulty level. When the validation loss for a specific task is higher, it indicates that this task has different patterns compared to the majority of tasks in $\mathcal{T}_{train}$ that dominates the training of the initial model. Hence, we can split the current set of training tasks $\mathcal{T}_{train}$ into Easy Task (E-$r$) and Hard Task (H-$r$), where $r=0,1,2,\ldots, u$ indicates the task layer (difficulty level of the easy-to-hard task hierarchy shown in Figure \ref{fig:framework}
(c)), by using a threshold $\gamma$ on the validation performance. Specifically, each task $T_i$ on the task layer $r$ is categorized as
\begin{equation}
D(T_i)=
 \begin{cases}
 \text{ Hard Task } (\text{H-}r) & \text{if } \text{R}^2 [ \mathcal{F}_{\theta_i}(x_q^{(i)}),y_q^{(i)}] < \gamma  \\ 
 \text{ Easy Task } (\text{E-}r) & \text{others }
\end{cases}
\label{eqa: gamma}
\end{equation}
where R$^2$ is the performance metric measured on the validation data of task $T_i$ from the Train Q-Set.
We repeat this process for every task and gather an array of R$^2$ for all the tasks. The threshold $\gamma$ is selected based on a statistical test over the obtained R$^2$ array, which will be discussed later.

\begin{algorithm}[t]
\SetAlgoLined
\DontPrintSemicolon
\SetKwInput{KwInput}{Input}
\SetKwInput{KwOutput}{Output}
\SetKwInput{KwLet}{Let}
\KwOutput{Optimized meta model weight $\theta^*$, Easy Task E-$r$, Hard Task H-$r$ in the task layer $r$}
\KwInput{Tasks $\mathcal{T}$, meta model $ \mathcal{F}_{\theta}$}
 Initialization: learning rate $\alpha, \beta$, task layer $r=0$ \;
 \While {not done} {
 \If{task layer r $\geq$ 1}{Current task $\mathcal{T}\leftarrow$ H-$r$}

 \While {not done} {
     \For{support tasks \(T_i\) in $\mathcal{T}$}{
        Adapt meta model \(\theta\) on task \(T_i\) by Eq. \ref{eq:theta_i}:
        \(\theta_i \leftarrow\theta -\alpha \nabla \mathcal{L}_{T_i} \left(\mathcal{F}_{\theta}\right)\) \;
    }

 Compute query loss $\mathcal{L}_{MAML}\left(\mathcal{F}_{\theta}\right)$ by Eq. \ref{eq:theta_loss} \;
 Update $\theta \leftarrow \theta-\beta \nabla_{\theta} \mathcal{L}_{TMAML}\left(\mathcal{F}_{\theta}\right)$ \;
   }
 Compute R$^2$, $\gamma$ using Alg \ref{algo:gamma} \;
 Split tasks $\mathcal{T}$ into E-$r$ and H-$r$ using $\gamma$ \;
 $r \leftarrow r + 1$
}
 \caption{Task-adaptive Meta Learning}
 \label{algo:adaptive-maml}
\end{algorithm}
\setlength{\textfloatsep}{3.2pt}
\begin{algorithm}[h!]
\SetAlgoLined
\DontPrintSemicolon
\SetKwInput{KwInput}{Input}
\SetKwInput{KwOutput}{Output}
\SetKwInput{KwLet}{Let}
\KwOutput{Threshold $\gamma$}
\KwInput{R$^2$ array; Lower/upper bounds ratio $a, b$}
 Initialization: Rank R$^2$ in ascending order; Set array $V=\emptyset$, $N$ the length of R$^2$ array, index $k=0$
 \;
 \While{$k$ is less than $N$}{
    $U$ $\leftarrow$ R$^2$[$:k$] \;
    $U'$ $\leftarrow$ R$^2$[$k:$] \;
    $V_k$ $\leftarrow$ Var($U$) + Var($U'$) \;
    $k \leftarrow k + 1$\;
 }
 $\gamma$ = R$^2[$ArgMin($V$[$\floor{aN}:\floor{bN}$])$]$
 \caption{Threshold $\gamma$ Selection Algorithm}
 \label{algo:gamma}
\end{algorithm}

\textbf{Training phase to build the hierarchy: } In the training phase, we iteratively bi-partition the set of hard tasks obtained from the previous task layer (H-$r$-$1$), where $r\geq1$.  The underlying intuition is to identify the set of tasks that cannot be well captured by the current model (H-$r$ Model). Starting from the second task layer, we build a meta initial model to be fine-tuned to the tasks in the current task set (i.e., all the  hard tasks from the previous task layer) via only a few gradient descent steps, following the standard MAML method. Again, we use the validation performance (measured by R$^2$) of each task-specific fine-tuned model to split the current task set into Easy Task (E-$r$) and Hard Task (H-$r$) sets via the threshold-based method (Eq. \ref{eqa: gamma}). Here the higher validation loss for a task indicates that the meta model cannot generalize well on this task with a small refinement. To expedite the training of the MAML method, we initialize the meta model with the predictive model (if $r=1$) or the meta model (if $r\geq2$) learned from the previous task layer. The process is summarized in Algorithm 1. 

\textbf{Selection  of split threshold: }
We discuss the selection of the threshold $\gamma$ for splitting the task set at each layer $r$ into Easy (E-$r$) and Hard (H-$r$) tasks. Given the obtained validation performance metrics $e_i$ (e.g., R$^2$) for each  task $i$, we aim to identify a subset of tasks that have significantly $\{e_i\}$ values compared to the remaining tasks. Hence, we adopt a statistical test, where the null hypothesis $H_0$ states that $e_i$ for all the tasks follow a single normal distribution while the alternative hypothesis $H_1$ states that there exists a subset of tasks $U$, and they follow a different normal distribution from the remaining tasks $U'$.  Here $U$ can be either the hard or easy tasks. The optimal set $U$ can be obtained by solving the following optimization problem: 
\begin{equation}
    U^*  = \text{argmax}_S  \log \frac{\text{Likelihood }(H_1|U)}{\text{Likelihood }(H_0)}.
\end{equation}
\noindent According to the prior work~\cite{xie2021statistically}, this can be solved by minimizing the sum of the variance of $U$ and $U'$. Hence we can select the threshold $\gamma$ that leads to the smallest value of the sum of Var$(U)$ and Var$(U')$. This process is summarized in Algorithm 2. 

\textbf{Testing phase using the hierarchy: }
In the testing phase, given any new task $T' \in \mathcal{T}_{test}$, we need to identify its difficulty level so that the learned meta models (E-$r$ Model and H-$r$ Model) on the corresponding task layer $r$ can be adapted to the new task. Specifically, starting from the initial model in the easy-to-hard task hierarchy shown in Figure \ref{fig:framework} (f), we adapt the corresponding meta model to the new task and measure its validation performance in the Test Q-Set. The obtained R$^2$ will be compared against the threshold $\gamma$ on the task layer $r$. Then we move to the next task layer based on the comparison outcome. This process is repeated until reaching a leaf node of the easy-to-hard task hierarchy.  Then we will adapt the final selected model (either the E-$r$ Model where $r=1,2,\ldots,u$ or the H-$r$ Model w.r.t. $r=u$) to the new task $T^\prime_i$ for testing. 
\section{Data and Experiments}

\setlength{\textfloatsep}{10pt}

\begin{figure*}[t]
    \centering
    \includegraphics[width=0.99\textwidth]{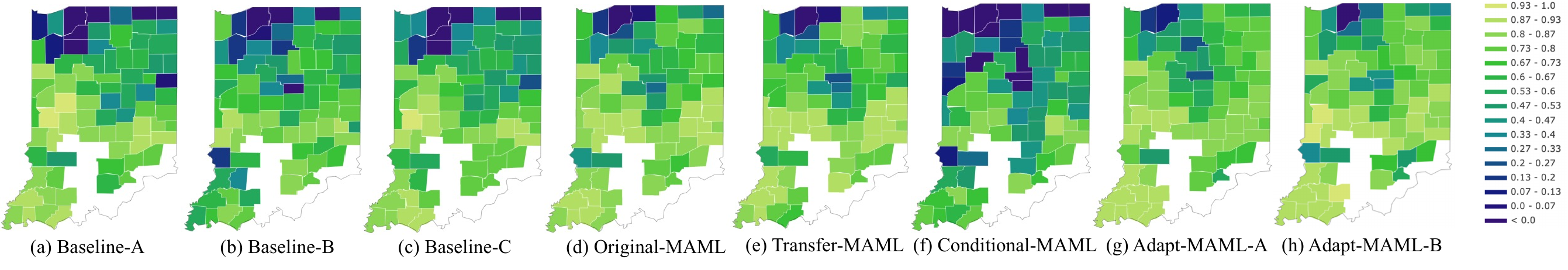}
    \vspace{-.12in}
    \caption{Spatially (county-level) visualized predicted R$^2$ performance in Indiana. The best model is the Adaptive-MAML-B.}
    \label{fig:test-visualization}
    \vspace{-.15in}
\end{figure*}

\subsection{Dataset}
The crop yield data are provided by USDA - National Agricultural Statistics Service (NASS) across 630 counties in the United States, and each county has 300 sampled datapoints. Each sample collects 19 daily features including weather and soil conditions, and plant properties, such as temperature, sand content, silt content, and crop yields in 21 consecutive years from 2000 to 2020 {\cite{liu2021estimating}}. Inspired by prior work~\cite{jia2021physics_tds,jia2021physics_simlr}, we generate the simulated data using a physics-based Ecosys model~\cite{zhou2021quantifying} over Illinois, Indiana and Iowa for pretraining the predictive model, which includes 10K county-level simulations with the same 19 daily features as NASS in 18 consecutive years. We define the  meta-learning tasks as county-level crop yield predictions thus we have 630 tasks in total. We construct the meta-learning dataset using the NASS data of which we sample 80\% counties in the Midwest of the United States, including Illinois, Wisconsin, Minnesota, Iowa, Missouri, Ohio, Michigan, North Dakota, South Dakota, Nebraska, Kansas, Kentucky, and Tennessee, as the training set and 20\% counties that are mostly in Indiana as the test set. In both training and testing sets, we use the first 5-year data from 2000 to 2004 as a support set, and the rest 16-year data from 2005 to 2020 as a query set. Next, we select 25 samples for every county in the support set as the Train Support Set (Train S-Set), and 75 samples for every county in the query set as the Train Query Set (Train Q-Set). We apply the same sampling strategy to the test set to obtain the Test Support Set (Test S-Set) and Test Query Set (Test Q-Set) sets. For the simulation data, we randomly sample 60\%, 20\%, and 20\% counties as synthetic training, validation, and test set, respectively.
 
\subsection{Candidate methods} We implement a diverse set of baselines and meta-learning-based models for model comparison.
\\
\textbf{Baseline-A} The predictive model trains on the Train S-Set and Train Q-Set, and tests on the Test Q-Set. \\
\textbf{Baseline-B} The predictive model trains on the Train S-Set, Train Q-Set and Test S-Set, and tests on the Test Q-Set. \\
\textbf{Baseline-C} The predictive model trains on the Train S-Set and Train Q-Set. Afterward, it fine-tunes on Test S-Set before testing on the Test Q-Set. \\
\textbf{Origin-MAML} The original MAML \cite{finn2017model} of which the meta model is trained on the Train S-Set and adapted on the Train Q-Set while training. In the test, the learned meta model is quickly fine-tuned on the Test S-Set before testing on the Test Q-Set. \\
\textbf{Transfer-MAML} The transfer meta-learning model \cite{soh2020meta}, which learns a global model on the Train S-Set, and then transfers the global weights to learn the MAML. \\ 
\textbf{Condition-MAML} The conditional meta-learning model \cite{denevi2020advantage}, which first trains several clusters (i.e., 4 clusters) using Train S-Set and learns the meta model to each cluster. In the test phase, it fine-tunes on the corresponding meta model based on its clustering prediction before testing.  \\
\textbf{Adaptive-MAML-A} The proposed adaptive meta-learning model implemented using Algorithm \ref{algo:adaptive-maml}. In this version, each Easy (E-$r$) and Hard (H-$r$) Model is trained with multiple inner epochs (i.e., 3 inner epochs), where  $r=1,2,\ldots,u$. \\
\textbf{Adaptive-MAML-B} The proposed adaptive meta-learning model that trains only 1 inner epoch on the Easy (E-$r$) and Hard (H-$r$) Model where $r=1,2,\ldots,u$. 
In the outer epoch iteration shown in Figure \ref{fig:framework} (e), it uses the H-$r$ w.r.t. $r=u$ (the hard model in the last task layer) learned in the previous outer epoch as the initial model for the current epoch training, except the first outer epoch that initializes with the pretrained predictive model.

\subsection{Implementation details} We implement the proposed task-adaptive meta-learning framework based on the learn2learn backbone \cite{arnold2020learn2learn} with Pytorch\footnote{The code is available at \url{https://github.com/ZhexiongLiu/Task-Adaptive-Meta-Learning}.}. We pretrain the predictive model with the simulated dataset that achieves 0.9898 R$^2$. Afterward, the pretrained predictive model is used to initialize the initial meta model weights in the first task layer of the easy-to-hard hierarchy. We use Adam optimizer with 0.001 learning rate. In the training phase, we use 32 tasks (out of 630 tasks) as a batch to learn the meta model. In the test phase, we quickly fine-tune the learned meta model for every separate task on its support set and report the performance on its query set. We use the Mean Squared Error (MSE) as the loss function, and R$^2$ as the evaluation metric. We set hyper-parameter $a$ and $b$ in Algorithm \ref{algo:gamma} as 0.35 and 0.65, respectively, which are experimental values that would guide the model to learn a good threshold ($\gamma$) through a subset of the tasks that are between these two bounds. If not specified, the default adaption number is 1, the (inner) epoch number is 1, and the maximum splitting number is 3.  We run 30 (outer) epochs with an Nvidia Titan X GPU and report the best performance.

\begin{figure*}[t]
    \centering
    \includegraphics[width=0.99\textwidth]{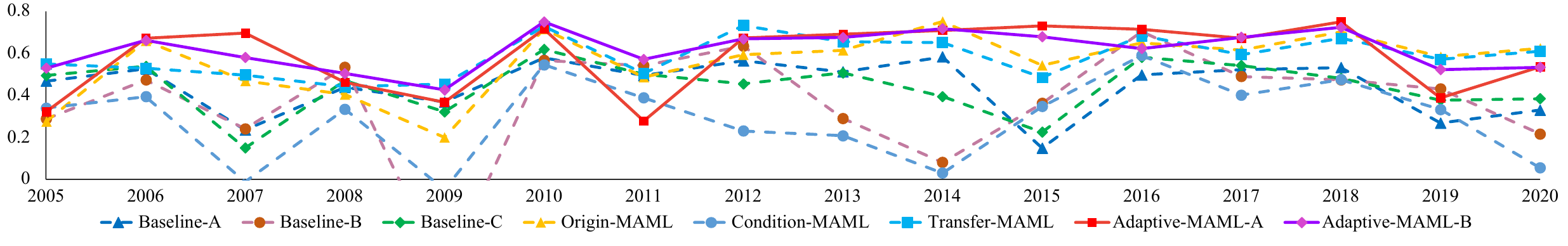}
    \vspace{-.1in}
    \caption{Temporally visualized predicted R$^2$ performance in 16 years from 2005 to 2020. The proposed Adaptive-MAMLs perform better than the other models in most years.}
    \label{fig:test-time-series}
    \vspace{-.06in}
\end{figure*}

\begin{figure*}[h!]
    \centering
    \includegraphics[width=0.99\textwidth]{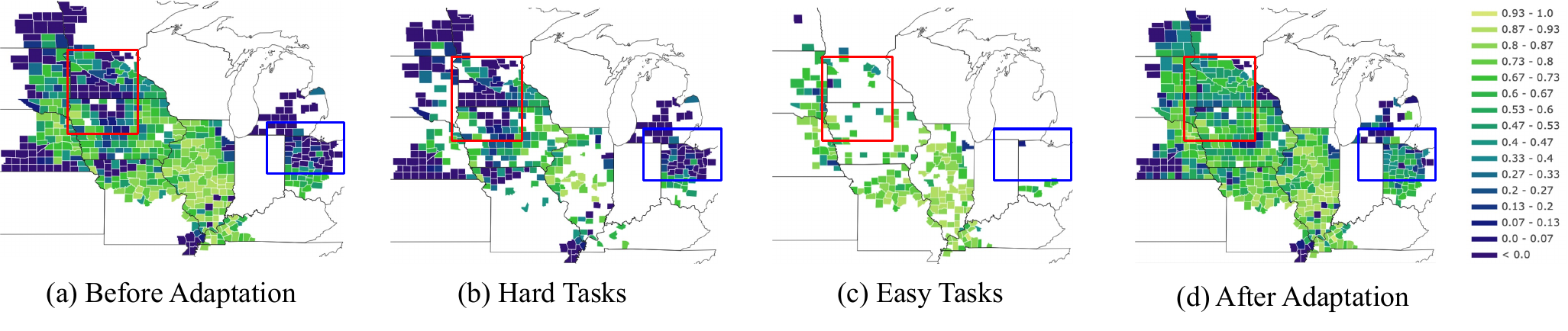}
    \vspace{-.15in}
    \caption{An example shows county-level R$^2$ improvement in (d) before and after adaptively training on (b) hard and (c) easy tasks, of which the areas enclosed by the red and blue boxes in (a) have low R$^2$.}
    \label{fig:adapt-before-after}
    \vspace{-.25in}
\end{figure*}

\begin{figure}[]
    \centering
    \includegraphics[width=0.73\columnwidth]{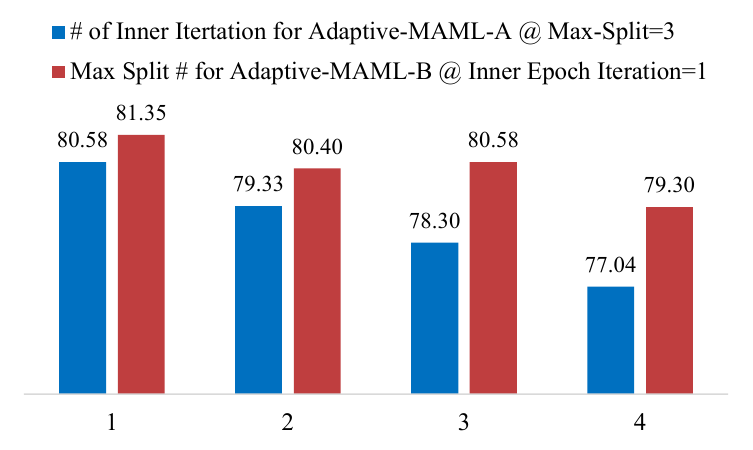}
    \vspace{-.1in}
    \caption{Parameter  sensitivity (with values ranging from 1 to 4) in terms of inner epoch iterations and the maximum split towards R$^2$ (\%) for the Adaptive-MAMLs.}
    \label{fig:parameter-sensitivity}
    \vspace{-.1in}
\end{figure}
\section{Analysis}

\begin{table}[t]
\caption{The MAML and baseline performance in terms of whole-yield, low-yield and high-yield predictions.}
\begin{adjustbox}{width=1\columnwidth}
\begin{tabular}{lccc}
\hline
\multirow{2}{*}{Models} & \multicolumn{3}{c}{R$^{2}$ (\%)} \\ \cline{2-4} 
 & Whole Yield & Low Yield & High Yield \\ \hline
Baseline-A & 71.34 & 37.96 & 62.83 \\
Baseline-B & 69.23 & 55.33 & 58.28 \\
Baseline-C & 70.62 & 36.76 & 63.34 \\
Origin-MAML & 78.98 & 60.62 & 73.07 \\
Condition-MAML & 62.84 & 35.81 & 54.35 \\
Transfer-MAML & 79.01 & 60.42 & 72.32 \\
Adaptive-MAML-A & 79.77 & 63.81 & 73.08 \\
Adaptive-MAML-B & \textbf{80.58} & \textbf{64.39} & \textbf{74.68} \\ \hline
\end{tabular}
\label{tab:main_results}
\end{adjustbox}
\end{table}

\subsection{Crop Yield Prediction}
In Table \ref{tab:main_results}, we compared the R$^2$ performance of different methods for the crop (corn) yield prediction. We consider the performance under three different sets of counties: (1) all the counties, (2) low-yield counties defined as the counties except the top 1/3 high-yield counties, and (3) high-yield counties that exclude the bottom 1/3 low-yield counties. As observed, our proposed adaptive-MAMLs outperform baselines and state-of-the-art meta-learning frameworks. In particular, the Adaptive-MAML-B achieves the best performance in all three sets. In the low-yield counties, the adaptive-MAMLs strongly dominate the baselines and state-of-the-art MAMLs. This is because the low-yield counties are usually along the state boundaries (as shown in Figure \ref{fig:example}) and are more difficult to model with existing machine learning methods. Hence, adaptively training on the hard tasks will help learn more on these poorly-performed counties. In terms of high-yield counties, they usually have lower data heterogeneity due to the technical improvement in crop cultivation thus the performance is relatively higher than the low-yield. In addition, Table \ref{tab:main_results_2} shows the MAML performance with 2 adaptations on trained meta models. As exhibited, the Adaptive-MAML-B improves the performance on the low-yield counties, which suggests that properly fine-tuning adaptation iterations will benefit low-yield counties. Moreover, the Adaptive-MAML-A performance is degraded (compared to Table \ref{tab:main_results}) because it is overfitting due to a large number of iterations (i.e., 2 adaptations in 3 inner epochs). 

\textbf{Spatial-dimension Performance} Figure \ref{fig:test-visualization} shows county-level predicted performance in Indiana. The Adaptive-MAML-B has mostly higher R$^2$ (light green in the figures) than the other models. In the baseline models, a region of the difficult tasks (counties) in the upper left boundary shows poor performance; however, our proposed adaptive-MAML greatly reduces the poorly-performed area, which indicates that our model is able to adaptively learn difficult tasks regardless of the spatial-related data heterogeneity. This is because our proposed framework learns a set of meta models that can be quickly adapted to different difficulty levels of tasks in the easy-to-hard hierarchy. 

\textbf{Temporal-dimension Performance} Figure \ref{fig:test-time-series} shows predicted R$^2$ performance in 16 consecutive years from 2005 to 2020. The Adaptive-MAML-A and Adaptive-MAML-B perform better than the other models in most of the year, but they perform relatively worse in the year of 2012. This is because the national-wide low crop yield occurs due to extreme weather and natural conditions. As for the year of 2019 and 2020, Adaptive-MAMLs have relatively poor performance because they are only designed to learn spatial heterogeneity thus may have limitations on capturing temporal data variability. This will be kept as our future work.

\begin{table}[t]
\caption{The MAML performance with 2 adaptations in terms of whole-yield, low-yield, and high-yield predictions.}
\begin{adjustbox}{width=1\columnwidth}
\begin{tabular}{lccc}
\hline
\multirow{2}{*}{Models} & \multicolumn{3}{c}{R$^{2}$ (\%)} \\ \cline{2-4} 
 & Whole Yield & Low Yield & High Yield \\ \hline
Origin-MAML & 78.53 & 61.90 & 71.19 \\
Condition-MAML & 65.28 & 41.80 & 57.93 \\
Transfer-MAML & 78.95 & 58.46 & 72.69 \\
Adaptive-MAML-A & 77.98    & 62.48 & 71.09 \\
Adaptive-MAML-B & \textbf{80.57} & \textbf{68.52} & \textbf{73.89} \\ \hline
\end{tabular}
\label{tab:main_results_2}
\end{adjustbox}
\end{table}

\begin{table}[t]
\centering
\caption{The splitting threshold $\gamma$ dynamically updates in each training epoch of the Adaptive-MAML-B model. The columns represent the threshold values for different task layers $r$ w.r.t. $r=1,2,3$ in the first 5 runs.}
\begin{adjustbox}{width=0.62\columnwidth}
\begin{tabular}{cccc}
\hline
$\gamma$ (\%) & Split-1 & Split-2 & Split-3 \\
\hline
Epoch1 & 40.56 & 32.68 & 16.47 \\
Epoch2 & 51.62 & 34.90 & 29.19 \\
Epoch3 & 58.35 & 48.63 & 46.12 \\
Epoch4 & 68.53 & 62.70 & 54.09 \\
Epoch5 & 73.48 & 67.41 & 61.05 \\\hline
\end{tabular}
\end{adjustbox}
    \label{tab:split-threshold}
\end{table}

\subsection{Parameter Sensitivity}
In this section, we discuss the parameter sensitivity in adaptive-MAMLs.
As shown in Figure \ref{fig:framework} (e) and (f), the splitting threshold $\gamma$ is updated each time either starting a new split or training a new epoch. We show the dynamic update of the threshold $\gamma$ in the first 5 epoch runs in Table \ref{tab:split-threshold}. As exhibited, the threshold $\gamma$ becomes higher as the number of epochs increases. During the training process, the adpative-MAML model is gradually updated to better fit training samples, and the $\gamma$ becomes higher as the model is setting a higher standard for hard tasks. In addition, the threshold $\gamma$ decreases with the splits going deeper of the easy-to-hard hierarchy, which indicates the tasks are more difficult in bottom task layers (e.g., $r=3$), thus the $\gamma$ becomes lower. In addition, we test different settings of inner epochs for Adaptive-MAML-A and the maximum split for Adaptive-MAML-B. The results (Figure \ref{fig:parameter-sensitivity}) show that the performance decreases given the increase of the inner epoch training. This shows that over-training under a premature task hierarchy can degrade the model performance. As for the maximum split, both 1, 2, and 3 are acceptable numbers. The small numbers mean fewer easy and hard models will be trained, which is helpful when the data has low heterogeneity.

\subsection{Case Study}
In Figure \ref{fig:adapt-before-after}, we study examples to show the effectiveness of adaptive-MAMLs. The areas (shown as counties on the map) marked with the red and blue rectangles in (a) exhibit low R$^2$ before adaptive learning. However, the Algorithm \ref{algo:gamma} splits current training tasks in (a) into (b) hard tasks and (c) easy tasks, and the Algorithm \ref{algo:adaptive-maml} iteratively trains on the hard tasks until a maximum splitting number is reached. As exhibited in (d), the poorly-performed areas (hard tasks) were greatly improved while well-performed areas (easy tasks) maintained excellence compared to (a). 

\section{Conclusion}
Standard meta-learning methods, e.g., MAML, can have degraded performance given a large number of heterogeneous tasks because spatial data variability is one of the most common issues in many spatial datasets. To bridge the gap, we proposed a task-adaptive MAML to learn spatial-related tasks with an easy-to-hard hierarchy that helps smoothly adapt the meta model to  new tasks. Extensive experiments show that our methods are superior to a diverse set of baselines and state-of-the-art models on real crop yield data in the Midwest of the United States. The proposed framework demonstrates meta-learning generalizability on a substantial number of spatial-sensitive meta models. In future work, we plan to develop robust MAMLs that are able to adaptively learn both spatial and temporal data heterogeneity, and ultimately promote the model feasibility to a wide range of  complex societal problems.

\section*{Acknowledgements}
This work was supported  by NSF awards 2147195, 2105133, and 2126474, NASA award 80NSSC22K1164,  the USGS awards G21AC10207, G21AC10564, and G22AC00266,  Google's AI for Social Good Impact Scholars program, the DRI award at the University of Maryland, and CRC at the University of Pittsburgh.

\bibliography{aaai23}

\begin{thebibliography}{49}
\providecommand{\natexlab}[1]{#1}

\bibitem[{Antoniou, Edwards, and Storkey(2019)}]{antoniou2018train}
Antoniou, A.; Edwards, H.; and Storkey, A. 2019.
\newblock How to train your MAML.
\newblock In \emph{International Conference on Learning Representations}.

\bibitem[{Arnold et~al.(2020)Arnold, Mahajan, Datta, Bunner, and
  Zarkias}]{arnold2020learn2learn}
Arnold, S. M.~R.; Mahajan, P.; Datta, D.; Bunner, I.; and Zarkias, K.~S. 2020.
\newblock learn2learn: {A} Library for Meta-Learning Research.
\newblock \emph{CoRR}, abs/2008.12284.

\bibitem[{Caruana(1997)}]{caruana1997multitask}
Caruana, R. 1997.
\newblock Multitask learning.
\newblock \emph{Machine learning}, 28(1): 41--75.

\bibitem[{Denevi, Pontil, and Ciliberto(2020)}]{denevi2020advantage}
Denevi, G.; Pontil, M.; and Ciliberto, C. 2020.
\newblock The advantage of conditional meta-learning for biased regularization
  and fine tuning.
\newblock \emph{Advances in Neural Information Processing Systems}, 33:
  964--974.

\bibitem[{Dong et~al.(2015)Dong, Wu, He, Yu, and Wang}]{dong2015multi}
Dong, D.; Wu, H.; He, W.; Yu, D.; and Wang, H. 2015.
\newblock Multi-task learning for multiple language translation.
\newblock In \emph{Proceedings of the 53rd Annual Meeting of the Association
  for Computational Linguistics and the 7th International Joint Conference on
  Natural Language Processing (Volume 1: Long Papers)}, 1723--1732.

\bibitem[{Elavarasan and Vincent(2020)}]{elavarasan2020crop}
Elavarasan, D.; and Vincent, P.~D. 2020.
\newblock Crop yield prediction using deep reinforcement learning model for
  sustainable agrarian applications.
\newblock \emph{IEEE access}, 8: 86886--86901.

\bibitem[{Finn, Abbeel, and Levine(2017)}]{finn2017model}
Finn, C.; Abbeel, P.; and Levine, S. 2017.
\newblock Model-agnostic meta-learning for fast adaptation of deep networks.
\newblock In \emph{International conference on machine learning}, 1126--1135.
  PMLR.

\bibitem[{Huang et~al.(2018)Huang, Zhang, Zheng, and
  Chawla}]{huang2018deepcrime}
Huang, C.; Zhang, J.; Zheng, Y.; and Chawla, N.~V. 2018.
\newblock DeepCrime: Attentive hierarchical recurrent networks for crime
  prediction.
\newblock In \emph{Proceedings of the 27th ACM international conference on
  information and knowledge management}, 1423--1432.

\bibitem[{Huang et~al.(2020)Huang, Huang, Liu, Dai, and Kong}]{huang2020lsgcn}
Huang, R.; Huang, C.; Liu, Y.; Dai, G.; and Kong, W. 2020.
\newblock LSGCN: Long Short-Term Traffic Prediction with Graph Convolutional
  Networks.
\newblock In \emph{IJCAI}, 2355--2361.

\bibitem[{Jia et~al.(2021{\natexlab{a}})Jia, Willard, Karpatne, Read, Zwart,
  Steinbach, and Kumar}]{jia2021physics_tds}
Jia, X.; Willard, J.; Karpatne, A.; Read, J.~S.; Zwart, J.~A.; Steinbach, M.;
  and Kumar, V. 2021{\natexlab{a}}.
\newblock Physics-guided machine learning for scientific discovery: An
  application in simulating lake temperature profiles.
\newblock \emph{ACM/IMS Transactions on Data Science}, 2(3): 1--26.

\bibitem[{Jia et~al.(2021{\natexlab{b}})Jia, Xie, Li, Chen, Zwart, Sadler,
  Appling, Oliver, and Read}]{jia2021physics_simlr}
Jia, X.; Xie, Y.; Li, S.; Chen, S.; Zwart, J.; Sadler, J.; Appling, A.; Oliver,
  S.; and Read, J. 2021{\natexlab{b}}.
\newblock Physics-Guided Machine Learning from Simulation Data: An Application
  in Modeling Lake and River Systems.
\newblock In \emph{2021 IEEE International Conference on Data Mining (ICDM)},
  270--279. IEEE.

\bibitem[{Jiang et~al.(2022)Jiang, Huang, Segovia-Dominguez, Newlands, and
  Gel}]{jiang2022learning}
Jiang, T.; Huang, M.; Segovia-Dominguez, I.; Newlands, N.; and Gel, Y.~R. 2022.
\newblock Learning Space-Time Crop Yield Patterns with Zigzag Persistence-Based
  LSTM: Toward More Reliable Digital Agriculture Insurance.
\newblock In \emph{Proceedings of the AAAI Conference on Artificial
  Intelligence}, volume~36, 12538--12544.

\bibitem[{Jiang et~al.(2020)Jiang, Huang, Geng, and Deng}]{jiang2020multi}
Jiang, W.; Huang, K.; Geng, J.; and Deng, X. 2020.
\newblock Multi-scale metric learning for few-shot learning.
\newblock \emph{IEEE Transactions on Circuits and Systems for Video
  Technology}, 31(3): 1091--1102.

\bibitem[{Karpatne et~al.(2018)Karpatne, Ebert-Uphoff, Ravela, Babaie, and
  Kumar}]{karpatne2018machine}
Karpatne, A.; Ebert-Uphoff, I.; Ravela, S.; Babaie, H.~A.; and Kumar, V. 2018.
\newblock Machine learning for the geosciences: Challenges and opportunities.
\newblock \emph{IEEE Transactions on Knowledge and Data Engineering}, 31(8):
  1544--1554.

\bibitem[{Kendall, Gal, and Cipolla(2018)}]{kendall2018multi}
Kendall, A.; Gal, Y.; and Cipolla, R. 2018.
\newblock Multi-task learning using uncertainty to weigh losses for scene
  geometry and semantics.
\newblock In \emph{Proceedings of the IEEE conference on computer vision and
  pattern recognition}, 7482--7491.

\bibitem[{Li, Zhang, and Huang(2020)}]{li2020learning}
Li, D.; Zhang, J.; and Huang, K. 2020.
\newblock Learning to learn cropping models for different aspect ratio
  requirements.
\newblock In \emph{Proceedings of the IEEE/CVF Conference on Computer Vision
  and Pattern Recognition}, 12685--12694.

\bibitem[{Li et~al.(2016)Li, Zhao, Wei, Yang, Wu, Zhuang, Ling, and
  Wang}]{li2016deepsaliency}
Li, X.; Zhao, L.; Wei, L.; Yang, M.-H.; Wu, F.; Zhuang, Y.; Ling, H.; and Wang,
  J. 2016.
\newblock Deepsaliency: Multi-task deep neural network model for salient object
  detection.
\newblock \emph{IEEE transactions on image processing}, 25(8): 3919--3930.

\bibitem[{Li et~al.(2017)Li, Zhou, Chen, and Li}]{li2017meta}
Li, Z.; Zhou, F.; Chen, F.; and Li, H. 2017.
\newblock Meta-sgd: Learning to learn quickly for few-shot learning.
\newblock \emph{CoRR}, abs/1707.09835.

\bibitem[{Liu et~al.(2021)Liu, Zhou, Jin, Tang, Jia, Jiang, Guan, Peng, Xu,
  Yang et~al.}]{liu2021estimating}
Liu, L.; Zhou, W.; Jin, Z.; Tang, J.; Jia, X.; Jiang, C.; Guan, K.; Peng, B.;
  Xu, S.; Yang, Y.; et~al. 2021.
\newblock Estimating the Autotrophic and Heterotrophic Respiration in the US
  Crop Fields using Knowledge Guided Machine Learning.
\newblock In \emph{AGU Fall Meeting Abstracts}, volume 2021, B25O--13.

\bibitem[{Liu et~al.(2019)Liu, He, Chen, and Gao}]{liu2019multi}
Liu, X.; He, P.; Chen, W.; and Gao, J. 2019.
\newblock Multi-Task Deep Neural Networks for Natural Language Understanding.
\newblock In \emph{Proceedings of the 57th Annual Meeting of the Association
  for Computational Linguistics}, 4487--4496. Florence, Italy: Association for
  Computational Linguistics.

\bibitem[{Ma, Du, and Matusik(2020)}]{ma2020efficient}
Ma, P.; Du, T.; and Matusik, W. 2020.
\newblock Efficient continuous pareto exploration in multi-task learning.
\newblock In \emph{International Conference on Machine Learning}, 6522--6531.
  PMLR.

\bibitem[{Matsumi and Yamada(2021)}]{matsumi2021few}
Matsumi, S.; and Yamada, K. 2021.
\newblock Few-Shot Learning Based on Metric Learning Using Class Augmentation.
\newblock In \emph{2020 25th International Conference on Pattern Recognition
  (ICPR)}, 196--201. IEEE.

\bibitem[{Mishra et~al.(2018)Mishra, Rohaninejad, Chen, and
  Abbeel}]{mishra2017simple}
Mishra, N.; Rohaninejad, M.; Chen, X.; and Abbeel, P. 2018.
\newblock A simple neural attentive meta-learner.
\newblock In \emph{International Conference on Learning Representations}.

\bibitem[{Munkhdalai and Yu(2017)}]{munkhdalai2017meta}
Munkhdalai, T.; and Yu, H. 2017.
\newblock Meta networks.
\newblock In \emph{International Conference on Machine Learning}, 2554--2563.
  PMLR.

\bibitem[{Nevavuori, Narra, and Lipping(2019)}]{nevavuori2019crop}
Nevavuori, P.; Narra, N.; and Lipping, T. 2019.
\newblock Crop yield prediction with deep convolutional neural networks.
\newblock \emph{Computers and electronics in agriculture}, 163: 104859.

\bibitem[{Nigam et~al.(2019)Nigam, Garg, Agrawal, and Agrawal}]{nigam2019crop}
Nigam, A.; Garg, S.; Agrawal, A.; and Agrawal, P. 2019.
\newblock Crop yield prediction using machine learning algorithms.
\newblock In \emph{2019 Fifth International Conference on Image Information
  Processing (ICIIP)}, 125--130. IEEE.

\bibitem[{Pan et~al.(2020)Pan, Zhang, Liang, Zhang, Yu, Zhang, and
  Zheng}]{pan2020spatio}
Pan, Z.; Zhang, W.; Liang, Y.; Zhang, W.; Yu, Y.; Zhang, J.; and Zheng, Y.
  2020.
\newblock Spatio-temporal meta learning for urban traffic prediction.
\newblock \emph{IEEE Transactions on Knowledge and Data Engineering}.

\bibitem[{Qiao et~al.(2018)Qiao, Liu, Shen, and Yuille}]{qiao2018few}
Qiao, S.; Liu, C.; Shen, W.; and Yuille, A.~L. 2018.
\newblock Few-shot image recognition by predicting parameters from activations.
\newblock In \emph{Proceedings of the IEEE Conference on Computer Vision and
  Pattern Recognition}, 7229--7238.

\bibitem[{Santoro et~al.(2016)Santoro, Bartunov, Botvinick, Wierstra, and
  Lillicrap}]{santoro2016meta}
Santoro, A.; Bartunov, S.; Botvinick, M.; Wierstra, D.; and Lillicrap, T. 2016.
\newblock Meta-learning with memory-augmented neural networks.
\newblock In \emph{International conference on machine learning}, 1842--1850.
  PMLR.

\bibitem[{Sharma, Rai, and Krishnan(2020)}]{sharma2020wheat}
Sharma, S.; Rai, S.; and Krishnan, N.~C. 2020.
\newblock Wheat Crop Yield Prediction Using Deep {LSTM} Model.
\newblock \emph{CoRR}, abs/2011.01498.

\bibitem[{Shekhar et~al.(2003)Shekhar, Zhang, Huang, and
  Vatsavai}]{shekhar2003trends}
Shekhar, S.; Zhang, P.; Huang, Y.; and Vatsavai, R.~R. 2003.
\newblock Trends in spatial data mining.
\newblock \emph{Data mining: Next generation challenges and future directions},
  357--380.

\bibitem[{Snell, Swersky, and Zemel(2017)}]{snell2017prototypical}
Snell, J.; Swersky, K.; and Zemel, R. 2017.
\newblock Prototypical networks for few-shot learning.
\newblock \emph{Advances in neural information processing systems}, 30.

\bibitem[{Soh, Cho, and Cho(2020)}]{soh2020meta}
Soh, J.~W.; Cho, S.; and Cho, N.~I. 2020.
\newblock Meta-transfer learning for zero-shot super-resolution.
\newblock In \emph{Proceedings of the IEEE/CVF Conference on Computer Vision
  and Pattern Recognition}, 3516--3525.

\bibitem[{Sung et~al.(2018)Sung, Yang, Zhang, Xiang, Torr, and
  Hospedales}]{sung2018learning}
Sung, F.; Yang, Y.; Zhang, L.; Xiang, T.; Torr, P.~H.; and Hospedales, T.~M.
  2018.
\newblock Learning to compare: Relation network for few-shot learning.
\newblock In \emph{Proceedings of the IEEE conference on computer vision and
  pattern recognition}, 1199--1208.

\bibitem[{Thrun and Pratt(1998)}]{thrun1998learning}
Thrun, S.; and Pratt, L. 1998.
\newblock Learning to learn: Introduction and overview.
\newblock In \emph{Learning to learn}, 3--17. Springer.

\bibitem[{Tseng et~al.(2021)Tseng, Kerner, Nakalembe, and
  Becker-Reshef}]{tseng2021learning}
Tseng, G.; Kerner, H.; Nakalembe, C.; and Becker-Reshef, I. 2021.
\newblock Learning to predict crop type from heterogeneous sparse labels using
  meta-learning.
\newblock In \emph{Proceedings of the IEEE/CVF Conference on Computer Vision
  and Pattern Recognition}, 1111--1120.

\bibitem[{Vinyals et~al.(2016)Vinyals, Blundell, Lillicrap, Wierstra
  et~al.}]{vinyals2016matching}
Vinyals, O.; Blundell, C.; Lillicrap, T.; Wierstra, D.; et~al. 2016.
\newblock Matching networks for one shot learning.
\newblock \emph{Advances in neural information processing systems}, 29.

\bibitem[{Wang et~al.(2020)Wang, Yao, Kwok, and Ni}]{wang2020generalizing}
Wang, Y.; Yao, Q.; Kwok, J.~T.; and Ni, L.~M. 2020.
\newblock Generalizing from a few examples: A survey on few-shot learning.
\newblock \emph{ACM computing surveys (csur)}, 53(3): 1--34.

\bibitem[{Willard et~al.(2021)Willard, Read, Appling, Oliver, Jia, and
  Kumar}]{willard2021predicting}
Willard, J.~D.; Read, J.~S.; Appling, A.~P.; Oliver, S.~K.; Jia, X.; and Kumar,
  V. 2021.
\newblock Predicting Water Temperature Dynamics of Unmonitored Lakes With
  Meta-Transfer Learning.
\newblock \emph{Water Resources Research}, 57(7): e2021WR029579.

\bibitem[{Wu, Zhang, and R{\'e}(2020)}]{wu2020understanding}
Wu, S.; Zhang, H.~R.; and R{\'e}, C. 2020.
\newblock Understanding and improving information transfer in multi-task
  learning.

\bibitem[{Xie et~al.(2021)Xie, He, Jia, Bao, Zhou, Ghosh, and
  Ravirathinam}]{xie2021statistically}
Xie, Y.; He, E.; Jia, X.; Bao, H.; Zhou, X.; Ghosh, R.; and Ravirathinam, P.
  2021.
\newblock A statistically-guided deep network transformation and moderation
  framework for data with spatial heterogeneity.
\newblock In \emph{2021 IEEE International Conference on Data Mining (ICDM)},
  767--776. IEEE.

\bibitem[{Xu et~al.(2020)Xu, Zhu, Zhong, Lin, Xu, Jiang, Huang, Li, and
  Lin}]{xu2020deepcropmapping}
Xu, J.; Zhu, Y.; Zhong, R.; Lin, Z.; Xu, J.; Jiang, H.; Huang, J.; Li, H.; and
  Lin, T. 2020.
\newblock DeepCropMapping: A multi-temporal deep learning approach with
  improved spatial generalizability for dynamic corn and soybean mapping.
\newblock \emph{Remote Sensing of Environment}, 247: 111946.

\bibitem[{Zhang and Yang(2018)}]{zhang2018overview}
Zhang, Y.; and Yang, Q. 2018.
\newblock An overview of multi-task learning.
\newblock \emph{National Science Review}, 5(1): 30--43.

\bibitem[{Zhang et~al.(2014)Zhang, Luo, Loy, and Tang}]{zhang2014facial}
Zhang, Z.; Luo, P.; Loy, C.~C.; and Tang, X. 2014.
\newblock Facial landmark detection by deep multi-task learning.
\newblock In \emph{European conference on computer vision}, 94--108. Springer.

\bibitem[{Zhao et~al.(2020)Zhao, Stretcu, Smola, and
  Gordon}]{zhao2020efficient}
Zhao, H.; Stretcu, O.; Smola, A.~J.; and Gordon, G.~J. 2020.
\newblock Efficient multitask feature and relationship learning.
\newblock In \emph{Uncertainty in Artificial Intelligence}, 777--787. PMLR.

\bibitem[{Zhao et~al.(2021)Zhao, Zhong, Yang, Luo, Lin, Li, and
  Sebe}]{zhao2021learning}
Zhao, Y.; Zhong, Z.; Yang, F.; Luo, Z.; Lin, Y.; Li, S.; and Sebe, N. 2021.
\newblock Learning to generalize unseen domains via memory-based multi-source
  meta-learning for person re-identification.
\newblock In \emph{Proceedings of the IEEE/CVF Conference on Computer Vision
  and Pattern Recognition}, 6277--6286.

\bibitem[{Zheng et~al.(2020)Zheng, Fan, Wang, and Qi}]{zheng2020gman}
Zheng, C.; Fan, X.; Wang, C.; and Qi, J. 2020.
\newblock Gman: A graph multi-attention network for traffic prediction.
\newblock In \emph{Proceedings of the AAAI conference on artificial
  intelligence}, volume~34, 1234--1241.

\bibitem[{Zhou et~al.(2019)Zhou, Zeng, Zhou, Anastasopoulos, and
  Neubig}]{zhou2019improving}
Zhou, S.; Zeng, X.; Zhou, Y.; Anastasopoulos, A.; and Neubig, G. 2019.
\newblock Improving robustness of neural machine translation with multi-task
  learning.
\newblock In \emph{Proceedings of the Fourth Conference on Machine Translation
  (Volume 2: Shared Task Papers, Day 1)}, 565--571.

\bibitem[{Zhou et~al.(2021)Zhou, Guan, Peng, Tang, Jin, Jiang, Grant, and
  Mezbahuddin}]{zhou2021quantifying}
Zhou, W.; Guan, K.; Peng, B.; Tang, J.; Jin, Z.; Jiang, C.; Grant, R.; and
  Mezbahuddin, S. 2021.
\newblock Quantifying carbon budget, crop yields and their responses to
  environmental variability using the ecosys model for US Midwestern
  agroecosystems.
\newblock \emph{Agricultural and Forest Meteorology}, 307: 108521.

\end{thebibliography}
\end{document}